\documentclass[10pt, journal]{IEEEtran}
\usepackage{graphicx}
\usepackage{multirow}
\usepackage{times}
\usepackage{adjustbox}
\usepackage{amsmath}
\usepackage{amssymb}
\usepackage{array}
\usepackage{color}
\usepackage{subcaption}
\usepackage{mathtools}
\usepackage{url}
\usepackage{bm}
\usepackage{breqn}
\usepackage{chronosys}

\begin{document}
\title{Frequency Disentangled Residual Network}

\author{
Satya Rajendra Singh$^{*}$, Roshan Reddy Yedla$^{*}$, Shiv Ram Dubey$^{\$}$, \IEEEmembership{Member,~IEEE,} Rakesh Sanodiya, and \\ Wei-Ta Chu, \IEEEmembership{Senior Member,~IEEE}
\thanks{S.R. Singh, R.R. Yedla and R. Sanodiya are with the Indian Institute of Information Technology, Sri City, Chittoor-517646, A.P., INDIA (e-mail: satyarajendra.rs@iiits.in, roshanreddy.y17@iiits.in, rakesh.s@iiits.in). }
\thanks{S.R. Dubey is with the Indian Institute of Information Technology, Allahabad, Prayagraj-211015, U.P., INDIA (e-mail: srdubey@iiita.ac.in). }
\thanks{W.T. Chu is with the National Cheng Kung University, Taiwan (e-mail: wtchu@gs.ncku.edu.tw). }
\thanks{$^{*}$Both authors have contributed equally.}
\thanks{$^{\$}$Corresponding Author.}
}
\markboth{Frequency Disentangled Residual Network}
 {Singh \MakeLowercase{\textit{}}: Bare Demo of IEEEtran.cls for Journals}

\maketitle

\begin{abstract}
Residual networks (ResNets) have been utilized for various computer vision and image processing applications. The residual connection improves the training of the network with better gradient flow. A residual block consists of few convolutional layers having trainable parameters, which leads to overfitting. Moreover, the present residual networks are not able to utilize the high and low frequency information suitably, which also challenges the generalization capability of the network. In this paper, a frequency disentangled residual network (FDResNet) is proposed to tackle these issues. Specifically, FDResNet includes separate connections in the residual block for low and high frequency components, respectively. Basically, the proposed model disentangles the low and high frequency components to increase the generalization ability. Moreover, the computation of low and high frequency components using fixed filters further avoids the overfitting. The proposed model is tested on benchmark CIFAR10/100, Caltech and TinyImageNet datasets for image classification. The performance of the proposed model is also tested in image retrieval framework. It is noticed that the proposed model outperforms its counterpart residual model. The effect of kernel size and standard deviation is also evaluated. The impact of the frequency disentangling is also analyzed using saliency map.
\end{abstract}


\section{Introduction}
The advent of Convolution Neural Networks (CNN) \cite{alexnet} has revolutionized the field of Computer Vision and Image Processing. The CNN models are a special kind of deep neural networks designed to deal with image data by utilizing the power of deep learning \cite{liu2017survey}. 
The success of the CNN models is accompanied due to the availability of large scale datasets and the prodigious improvement in computing power over the years \cite{lecun2015deep}. 
Another important reason for the success of these models is that amount of pre-processing required to analyse images is much minimal compared to the traditional machine learning algorithms. 
The CNN models \cite{alexnet}, \cite{googlenet}, \cite{resnet} take images as the input and learns the important features automatically from data in a hierarchical manner in contrast to the hand engineered feature extraction was being used in early days \cite{iold}, \cite{lwp}, \cite{mdlbp}.

The CNN models process the image data through a hierarchy of linear and non-linear functions \cite{schmidhuber2015deep}. They are an amalgam of various convolution layers, pooling layers, batch-normalization layers, dropouts, etc. The CNN models have shown very propitious performance for different computer vision and image processing tasks like scene recognition \cite{liu2021scene}, image classification \cite{resnet}, \cite{basha2020impact}, \cite{dubey2019diffgrad}, object localization \cite{ren2015faster}, image segmentation \cite{he2017mask}, image compression \cite{ma2019iwave}, image restoration \cite{jin2019flexible}, face recognition \cite{schroff2015facenet}, \cite{srivastava2019performance}, visual relationship detection \cite{tajrobehkar2021align}, medical image analysis \cite{choi2020combining}, \cite{lbpdad}, image super-resolution \cite{tian2020coarse}, etc., to name a few. In the past decade, several CNN models have been investigated for different applications. This is a very active research area at present to explore the different ways of designing of CNN models in a quest for better accuracy on unseen data. 
In this paper, we improve the CNN model by incorporating the frequency disentangling in the network in terms of the different paths for the high and low frequency feature extraction. The proposed model is able to generalize the CNN model over test data with improved performance.

\subsection{Motivation}
The residual neural network (ResNet) \cite{resnet} is a widespread deep CNN architecture that has shown very appealing performance for different image processing tasks such as Rain Removal \cite{que2020attentive}, Image Super-resolution \cite{park2021dynamic}, Image Compression \cite{akbari2021learned}, Object Detection \cite{chen2020reverse}, \cite{zhou2020salient}, Pediatric MRI Quality Assessment \cite{liu2020real}, Blur Detection \cite{tang2020br}, Image Dehazing \cite{yeh2019multi}, and Facial Attractiveness Computation \cite{fan2017label}.
The ResNet model utilises the identity connection to enable the transfer of information across layers without attenuation, leading to smoother optimization. The ResNet primarily addresses the issue of vanishing gradients in CNN network as the identity connection provides a highway for gradient to flow during back-propagation.
Though there exists several CNN architectures studied over the years, the ability of the model to generalize over unseen data and be more robust is still a concern. Thus, it is an urgent need to explore and design sophisticated CNN models to improve the accuracy on unseen data. 
Some researchers have tried to improve the performance of the CNN models over unseen data. Most of the attempts made were around tweaking the network architecture, the way the data is processed by the network or adding different loss functions in order to achieve better accuracy. Major attempts include adding regularisation layers like batch normalisation \cite{batchnorm}, different data augmentation techniques \cite{perez2017effectiveness}, dropout layers \cite{dropout}, ensemble generalization loss \cite{choi2020combining}, etc. have been tried out to improve the performance. The effect of different aspects of CNN have been also analyzed by different researchers in the recent past, such as kernel size and number of filters of convolution layer \cite{agrawal2020using}, fully connected layers \cite{basha2020impact}, activation functions \cite{hayou2018selection}, loss functions \cite{srivastava2019performance}, etc.

While these works opened new doors toward the generalization of the CNN models, their generalization capability is limited. 
In a preliminary work in \cite{yedla2021performance}, the performance of the CNN models is tested on the high and low frequency components of the image which served as a motivation for us to include the frequency disentangling as part of the CNN architecture to improve the generalization capability and robustness of the CNN model. 
Another motivation to utilize the frequency disentangling is to upgrade the CNN model with human perception which works differently for high and low frequency information \cite{vuilleumier2003distinct}, \cite{monson2014perceptual}.
The disentangling of different characteristics has also improved performance for different applications such as scale-disentangling for representation learning and image synthesis using a branch generative adversarial network \cite{yi2020bsd}, continuous low-shot detection using disentangling-imprinting-distilling \cite{chen2020did}, disentangled cross-modal fusion for object detection \cite{chen2020rgbd}, and disentangling perceptual and noisy features for blind quality estimation \cite{kottayil2017blind}.
Hence, motivated by the success of disentangling to separate the different characteristics in the features, we propose the frequency disentangling.
In this paper, we incorporate the frequency disentangling in the residual block of ResNet model and observe a significant improvement in the performance. 
Based on the success of the proposed frequency disentangling idea, a new family of frequency disentangled CNN architectures may spark in future.

\begin{figure*}[!t]
    \centering
    \includegraphics[width=0.9\textwidth]{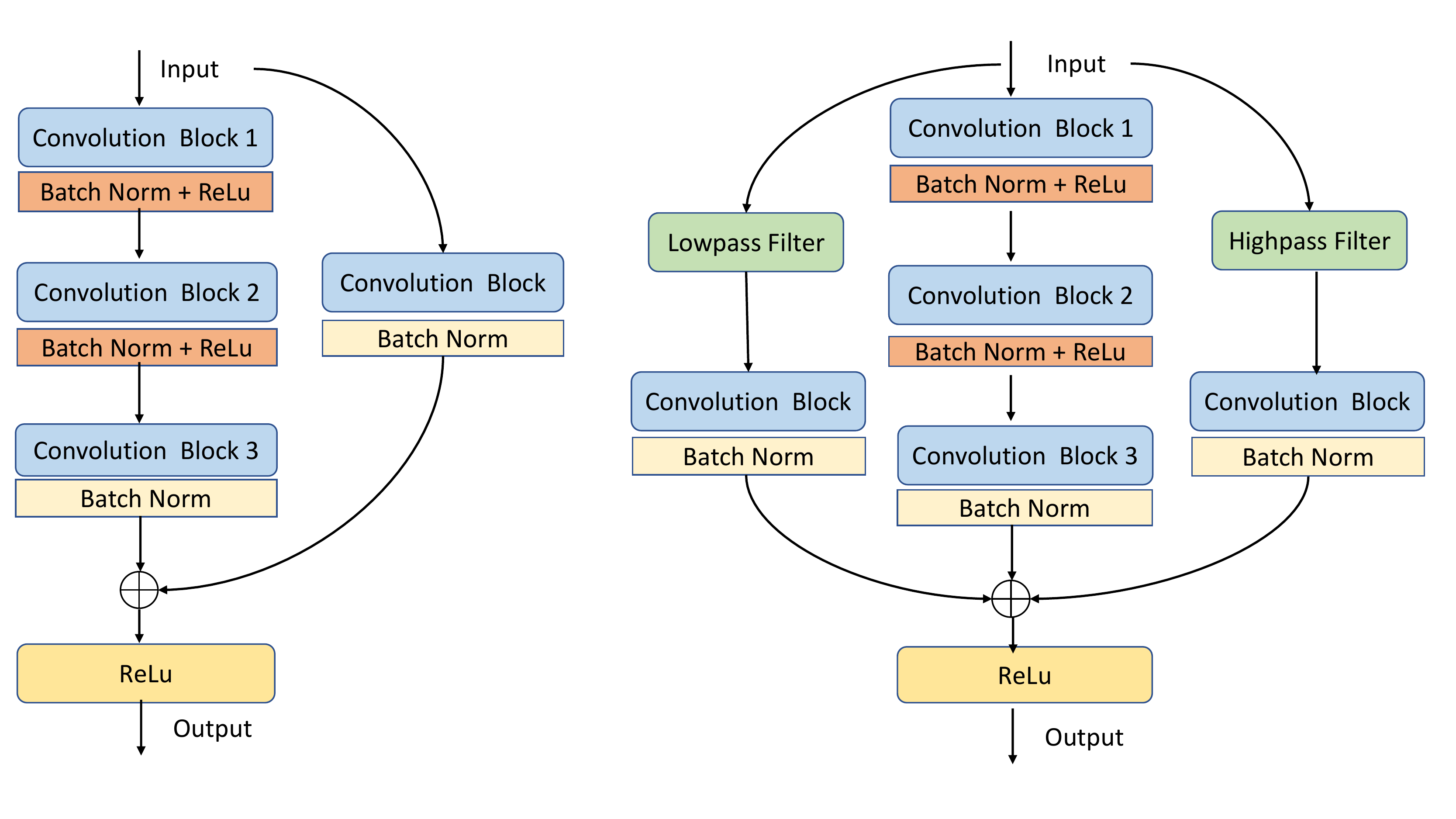}
    \caption{(\textit{left}) Existing bottleneck residual block of ResNet architectures \cite{resnet} and (\textit{right}) The proposed bottleneck block of FDResNet using two skip connections. Batch Norm refers to the batch normalization regularizer \cite{batchnorm} and ReLU refers to the rectified linear unit \cite{alexnet} activation function.}
    \label{fig:resnet}
\end{figure*}

\subsection{Related Works}
In recent years numerous CNN networks have been investigated. The networks have got deeper and deeper and evolved over time owing to the improved hardware, ImageNet competitions, and innovative ideas \cite{imagenet}. The AlexNet \cite{alexnet} was a revolutionary state-of-the-art CNN model winning the ImageNet Large-Scale Object Recognition Challenge in 2012 \cite{imagenet} achieving outstanding performance in contrast to the hand-designed methods. The AlexNet model utilized the ReLU activation function for better training.
Inspired by the success of the AlexNet, many new networks were built and studied. The VGGNet \cite{vggnet} is a popular family of CNN networks with more number of layers. The success of VGGNet model came from stacking uniform convolutions and building the deeper network. The GoogleNet \cite{googlenet} could solve this problem with its inception module. It does feature detection at different scales with different filters which reduced computation cost and improved accuracy.
Though, the models like AlexNet, VGGNet and GoogleNet were promising, but suffering from the training instability with deeper networks.

In a revolutionary work, the ResNet \cite{resnet} model tackles the training instability with the help of residual connections. Thus, the ResNet model has become the most successful CNN architecture in recent past. Specifically, the residual connection serves as skip connection which resolves the vanishing gradients issue as the gradient can flow back through the identity mapping. This innovative use of skip connections has led to the successful training of the deeper networks. The ResNet model has also served as the pioneers of using batch normalisation layers in the network for generalization.
The idea of skip connections has also been used in many other CNN architectures. The DenseNet \cite{densenet} is one such popular network. But unlike the ResNet where information is passed from one residual block to the next, in DenseNet, each layer gets input from all its preceding layers and provides its output to all the subsequent layers. All the inputs at each layer are concatenated, the intuition being each layer has a collective knowledge of all preceding ones.
the ResNeXt \cite{resnext} is another popular variant of the ResNet model. It has a number of bottleneck layers stacked in parallel fashion. The number of paths inside the ResNeXt block is termed as cardinality, usually set to 32. 
It showed that increasing the cardinality instead of depth and width is more favorable for reduced error. The ResNet with stochastic depth \cite{resnetstocasticdepth} comes with another novel idea. It tackles the long training time of the deep ResNet models and any possible overfitting with stochastic depth. 
In another attempt, the Squeeze and Excitation Network (SENet) \cite{senet} based ResNet re-configures the channels of the intermediate activation maps with the help of an excitation score. 
Most of the ResNet variants led to some improvement in the performance. Though the ResNet model has been extended by different means, the frequency disentangling is not much explored.

\subsection{Our Contributions}
In this paper, we harness the capability of frequency disentangling with ResNet model to improve the generalization ability. The major contributions of this paper are summarized as follows:
\begin{itemize}
\item We propose a frequency disentangled residual block by involving the additional high and low pass filtering modules.
\item We propose the frequency disentangled residual network (FDResNet) by stacking the frequency disentangled residual blocks.
\item The proposed FDResNet model is able to process the low and high frequency information separately, mimicking the human perception.
\item The proposed frequency disentangling concept leads to the better generalization capability of the model.
\item In order to show the importance of the proposed idea, we perform a rigorous experimental analysis for image classification and image retrieval on benchmark datasets. We also visualize the effect of FDResNet using saliency map.
\end{itemize}

\section{Proposed Frequency Disentangled Residual Network (FDResNet)}
In this section, first we provide an overview of ResNet then we present the proposed Frequency Disentangled Residual Network (FDResNet) model.

\subsection{ResNet Overview}
The Residual blocks are arguably one of the great advancements in the deep learning field in the last few years. ResNets have become a powerful architecture for training models for different applications over the last few years like Image classification, Object detection, Image retrieval, etc. The residual connection has become a major breakthrough and facilitated training deep networks possible with an improvement in the performance.  
A common practice in the development of CNN models is to employ a sequence of convolution layers along with other layers like batch normalization, activation function, pooling, etc. for building a deep CNN model. However, the plain models such AlexNet and VGGNet suffer due to the diminishing gradient during backpropagation of gradients in a deep network. The diminishing gradient kills the learning process and impacts the model's performance.

The ResNet cleverly tackles this problem by introducing the residual blocks as shown in left subfigure of Fig. \ref{fig:resnet}. The residual blocks have \emph{shortcut connections} which help to alleviate the problem of vanishing gradients. In a residual block the goal of the layers is not to directly fit the desired mapping of the data, but to fit a residual mapping. This in a sense allows for the actual data to be accessible in the deeper layers too. 
To make better sense of the residual, consider a neural network is being trained to map an input \(x\) to an output, which is a function of \(x\) as \(H(x)\). A residual is the difference between the input and output. In the residual block, the model tries to learn the residual from given \(x\) instead of directly calculating the \(H(x)\). The residual can be denoted as \(R(x)\). Thus residual of x can be represented as \(R(x) = H(x) - x\). As per observations, it is easier to learn this residual than the actual output. Moreover, the identity function in this setup can simply be got by setting residual to \(0\), making \(H(x) = x\). In this paper, we refer to the generalized ResNet which includes a Convolution layer and Batch Normalization layer along with the skip connection as shown in the left subfigure of Fig. \ref{fig:resnet}. Thus, the residual can be given as,
\begin{equation}
R(x) = H(x) - S(x)
\end{equation}
for generalized residual block where S(x) refers to the output of skip connection operated by convolution and batch normalization layers.
The residual block does not consider the information in a different frequency band, which is regarded as the major limitation and incorporated in the proposed model as detailed in the next subsection.  

\begin{figure*}[!t]
    \centering
    \includegraphics[width=0.9\textwidth]{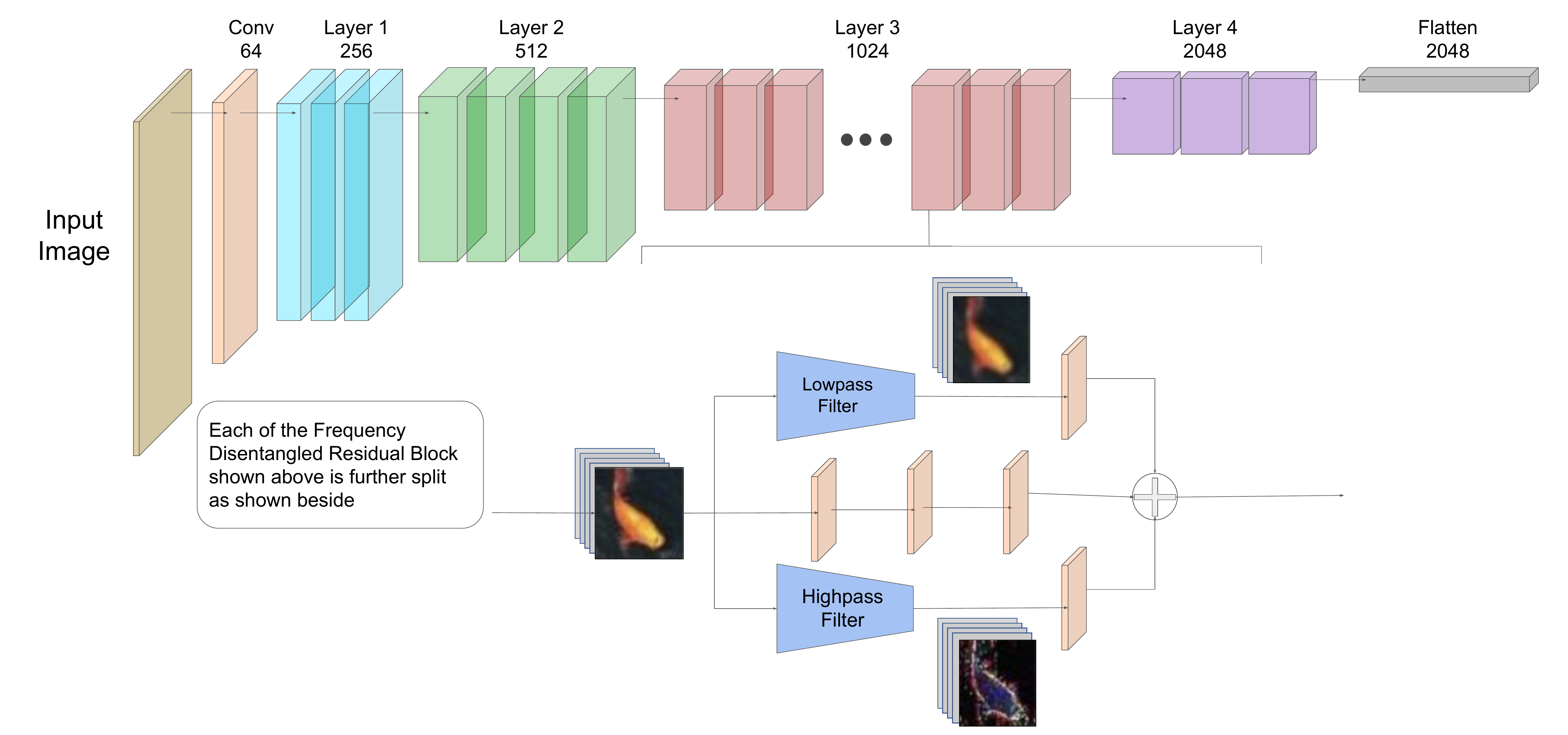}
    \caption{The proposed FDResNet architecture.}
    \label{fig:fdresnet}
\end{figure*}

\subsection{Proposed FDResNet}
The Residual blocks have been quite successful at tackling the problem of vanishing gradients and improving the performance of the CNN model. However, the ResNet model is not able to disentangle the high and low frequency information which leads to the final features with ambiguity resulting in poor generalization.
In order to tackle this problem, we propose a frequency disentangled residual network (FDResNet). 
The proposed FDResNet uses the frequency disentangled residual block which consists of low and high frequency disentangling as shown in right subfigure of Fig. \ref{fig:resnet}.
The proposed FDResNet has shown very promising performance for image classification and image retrieval on benchmark datasets like CIFAR, TinyImageNet, and Caltech-256.

As stated before the residual block consists of a skip connection, which carries the output of the previous block directly to the output of the current block. However, the frequency disentangled residual block in FDResNet contains \emph{two such skip connections} or residual connections. Moreover, these connections have an additional block that filters the image using a \emph{Gaussian kernel}. One of the skip connections contain a high pass filter and the other contains a low pass filter. So in this setup, the output of the previous layer is initially filtered in both of these skip connections where one acts as a high pass filter and the other as a low pass filter.
Consider $F_L$ and $F_H$ as the low pass and high pass filtering with Gaussian kernel having standard deviation $\sigma_L$ and $\sigma_H$, respectively. The residual in the proposed model can be given as,
\begin{equation}
    R(x) = H(x) - S(F_L(x,\sigma_L)) - S(F_H(x, \sigma_H)).
\end{equation}
Note that the kernels for high pass and low pass are considered independently. 
In the end, the output of the stacked layers, the high pass layer, and the low pass layers are added and operated through the non-linear activation function. Fig. \ref{fig:resnet} provides a side-by-side comparison between the residual block used in ResNet and the proposed frequency disentangled residual block used in FDResNet. The frequency disentangled residual blocks are stacked to form a deep FDResNet as depicted in Fig. \ref{fig:fdresnet}.

The high pass and low pass images are the results of image filtering with the corresponding kernel.
The high pass and low pass functions have specific kinds of kernel functions, which let only the high frequency or low frequency information in an image to appear in the output, respectively. The high frequency in an image usually contains the detail information such as edge, corner, etc. Thus, the high pass activation consists of prominent edge like information. Similarly, when only low frequency information is passed, the edges will be removed making the image smoother giving a blurred effect. 
In the proposed model, we add the high pass and low pass image filtering into the residual block which helps the model to process the low and high frequency of the input separately. It leads the proposed model to be more aligned with the human perception of dealing with the low and high frequency components separately and finally combine both the information. Thus, the proposed frequency disentangling based model learns data specific low and high frequency information as required and poses better generalization and robustness.

\subsection{Justification}
The residual block has a single skip connection which takes the entire input to the output of the stacked layers. The gradients in turn will also contain the information of the entire image as a whole. On the other hand in the proposed residual blocks of the FDResNet, the input is processed using low pass and high pass filtering before being fed to the output. Furthermore, the image's high frequency and low frequency information are added separately. 

We hypothesize that having these separate highways in the residual block allows the model to disentangle the frequency information. The model processes the low frequency and high frequency information separately. This helps the model to learn the data specific low and high frequency components. For example, there can be images with dominant low frequency information such as homogeneous regions and dominant high frequency information such as textured regions. Thus, in case of homogeneous regions the low pass skip connection will play the major role while in the case of textured regions the high pass skip connection will play the major role. 

Furthermore, since the outputs of the two skip connections are added back, there is no much loss of information. It can be understood that the model is simply able to draw out the coarser and finer details in an image at each layer before passing it to its subsequent layers. Also, each group of stacked layers may have different representations of the image as the input moves deeper. Thus, at all these layers, the model is able to look for fine distinctions by checking its low frequency and high frequency information, without much loss of information. 
The gradients during backpropagation will be also impacted with the high and low frequency operation, thus helping the model to improve the training for better generalization.
Having the separate skip connections for low frequency and high frequency helps the model to look at corresponding information in the image leading to more distinctive and robust feature learning.

\section{Experimental Setup}
In this section, we summarize the different datasets used, different networks utilized and training settings. We utilized different computational resources for the experiments including a system with 12 GB NVIDIA Titan XP GPU, a system with 24 GB NVIDIA RTX GPU and cloud based GPU service provided by Google Colab free tier. The experiments are carried out using the PyTorch deep learning library.

\begin{table*}[!t]
\caption{The classification accuracy using the ResNet model and the proposed FDResNet model on CIFAR-10, CIFAR-100, Caltech-256 and TinyImageNet datasets. In this experiment the sigma value is set as 1. Note that ResNet50 and FDResNet50 are used on CIFAR-10 and CIFAR-100 datasets. While ResNet101 and FDResNet101 are used on Caltech-256 and TinyImageNet datasets. The average accuracy over three trials is reported along with standard deviation. The best results on a dataset is highlighted in bold.}
\centering
\renewcommand{\arraystretch}{1.2}
\resizebox{\textwidth}{!}{
\begin{tabular}{|p{0.2\columnwidth}|p{0.2\columnwidth}|p{0.2\columnwidth}|p{0.2\columnwidth}|p{0.2\columnwidth}|p{0.2\columnwidth}|p{0.2\columnwidth}|}
\hline
\multirow{2}{*}{Dataset} & ResNet & \multicolumn{5}{c|}{FDResNet Accuracy using kernel with sigma 1} \\ 
 \cline{3-7} 
& Accuracy & (L3, H3) & (L5, H5) & (L7, H7) & (L3, H5) & (L5, H3) \\ \hline
CIFAR-10 &
  ${95.16\pm0.12}$ &
  ${\textbf{95.32}\pm0.16}$ &
  ${95.20\pm0.20}$ &
  ${95.13\pm0.25}$ &
  ${94.95\pm0.44}$ &
  ${95.29\pm0.24}$ \\ \hline
CIFAR-100 &
  ${78.42\pm0.60}$ &
  ${\textbf{80.45}\pm0.04}$ &
  ${79.91\pm0.14}$ &
  ${79.70\pm0.25}$ &
  ${79.45\pm0.42}$ &
  ${79.84\pm0.18}$ \\ \hline
Caltech-256 &
  ${58.17\pm0.73}$ &
  ${60.74\pm0.32}$ &
  ${61.51\pm2.13}$ &
  ${\textbf{62.82}\pm2.22}$ &
  ${55.55\pm0.48}$ &
  ${62.37\pm2.65}$ \\ \hline
TinyImageNet &
  ${57.18\pm0.35}$ &
  ${61.97\pm0.27}$ &
  ${\textbf{62.47}\pm0.26}$ &
  ${62.08\pm0.08}$ &
  ${57.42\pm0.72}$ &
  ${60.76\pm0.62}$ \\ \hline
\end{tabular}
}
\label{tab:classification_results}
\end{table*}

\subsection{Datasets Used}
In order to justify our findings that the proposed FDResNet is superior to the ResNet, we test the models on four widely used image classification datasets, including CIFAR-10 \cite{krizhevsky2009learning}, CIFAR-100 \cite{krizhevsky2009learning},  Caltech-256 \cite{caltech256}, and TinyImageNet \cite{le2015tiny}.
The CIFAR-10 dataset\footnote{https://www.cs.toronto.edu/~kriz/cifar.html} contains $50,000$ training images and $10,000$ testing images which are equally distributed in $10$ classes. Thus, each class contains $5,000$ images for training and $1,000$ images for testing, making a total of $60,000$ images in the dataset. 
The CIFAR-100 dataset contains similar images as CIFAR-10 dataset. However, it contains $100$ classes with each having $500$ training images and $100$ testing images. The CIFAR-100 is more fine-grained as compared to the CIFAR-10. The dimension of the images in CIFAR-10 and CIFAR-100 datasets is $32 \times 32 \times 3$. 
The Caltech-256 dataset\footnote{http://www.vision.caltech.edu/Image_Datasets/Caltech256/} contains $30,607$ images distributed unequally in $257$ diverse categories. One class is a clutter class containing random images not belonging to any of the other $256$ classes and ignored for training. Thus, the total number of images from $256$ category is $29,780$. Around $15\%$ of images from each class are taken randomly for the testing set having total $4,841$ images and remaining $24,939$ images for the training set. Note that the dimension of images varies in Caltech-256 dataset.
The TinyImageNet dataset\footnote{https://www.kaggle.com/c/tiny-imagenet} contains $1,00,000$ images for training and $10,000$ images for testing, equally divided in $200$ classes. Thus, each class contains $500$ images for training and $50$ images for testing. The dimension of the images in TinnyImagenet dataset is $64 \times 64 \times 3$.

\subsection{Networks Used}
Since the proposed FDResNet model is primarily based on the ResNet architecture, we compare the results of FDResNet with ResNet model of different depths. We use the FDResNet50 with different combinations of sigma (i.e., the standard deviation of Gaussian kernel) and kernel size for the filtering blocks in both low-pass and high-pass skip connections. We also test the proposed model with only either low-pass or high-pass skip connection. We use ResNet50 and FDResNet50 on CIFAR10 and CIFAR100 datasets and ResNet101 and FDResNet101 on Caltech-256 and TinyImageNet datasets.

\subsection{Experimental Settings}
The training setting varies for different experiments. Firstly for data augmentation, the images of CIFAR10, CIFAR100 and TinyImageNet datasets are used in the same sizes, i.e., $32\times32$, $32\times32$ and $64\times64$, respectively. However, the images of Caltech-256 dataset are resized to $128\times128$. We also use random flip and normalization on these images. We use Cross Entropy Loss and Stochastic Gradient Descent optimization with $0.9$ momentum and $5e^{-4}$ weight decay for image classification experiments. 
The ResNet50 and FDResNet50 are trained for $200$ epochs on CIFAR-10 and CIFAR-100 datasets. While, the ResNet101 and FDResNet101 are trained for $80$ epochs on Caltech and TinyImageNet datasets. The multi-step learning rate scheduler is also utilized in all the experiments. In the case of ResNet50 and FDResNet50, the learning rate initially starts at $0.1$ and drops by a factor of $0.1$ at $120^{th}$ and $170^{th}$ epoch. For Caltech-256 and TinyImageNet, the learning rate starts at $0.1$ and drops by a factor of $0.1$ at every $20^{th}$ epoch.
We calculate the classification accuracy by the proposed FDResNet with different settings for each dataset and compare it with the accuracy by ResNet. It is to be noted that all the models are trained three times with independent initialization in each experiment to ensure the consistency of the results. The average over three trails along with standard deviation is reported.
We experiment with $0.5$, $1$, and $1.5$ values of sigma (i.e., standard deviation) of Gaussian filtering. We also test the trainable sigma for the low-pass and high-pass filtering. The kernel size is considered as $3$, $5$ and $7$. 

In most of the experiments, we use the sigma value as $1$ in the filtering blocks of the proposed FDResNet. With sigma value as $1$, we perform $9$ experiments on each dataset with the different combination of kernel sizes of the filtering block. The kernel sizes for high-pass skip connection and low-pass skip connection are considered as (L3, H3), (L5, H5), (L7, H7), (L3, H5), (L5, H3), (L3, Nil), (Nil, H3), (L5, Nil), and (Nil, H5)\footnote[1]{In this notation used here as well as at other places, the $1^{st}$ value in the brackets is the kernel size for high-pass filtering and the $2^{nd}$ value is the kernel size for low-pass filtering.}. Note that Nil refers to the scenario where only one skip connection (i.e., either low-pass or high-pass) is used and the skip connection corresponding to Nil is missing. We also test the suitability of the proposed model with $0.5$, $1.5$ and trainable sigma of Gaussian filtering with (L3, H3), (L5, H5), and (L7, H7) kernel sizes for high-pass and low-pass filtering. The results and analysis for each of these experiments are presented in the next section.

\begin{figure*}[!t]
 \begin{subfigure}{\textwidth}
   \centering
\includegraphics[width=1\textwidth]{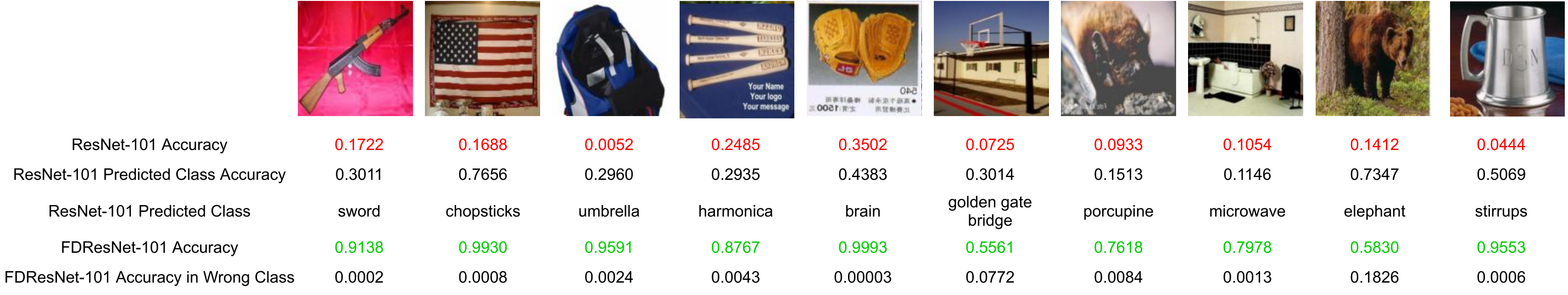} 
\caption{The prediction results using ResNet101 and FDResNet101 with a sigma value of 1 and kernel size of 3 for low pass skip connection and 5 for high pass skip connection on sample Caltech-256 images. }
\label{fig:Caltech_Scores}
\end{subfigure}
\\[2ex]
\begin{subfigure}{\textwidth}
\centering
 \includegraphics[width=1\textwidth]{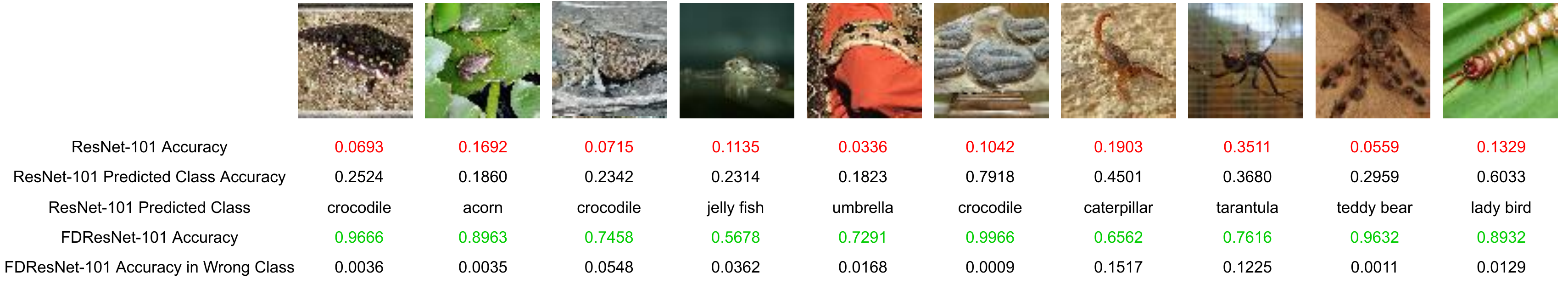} 
\caption{The prediction results using ResNet101 and FDResNet101 with a sigma value of 1 and kernel size of 5 for both low pass and high pass skip connections on sample TinyImageNet images. }
\label{fig:Tiny_Scores} 
\end{subfigure}
\caption{In both these figures, \emph{ResNet-101 accuracy} is accuracy of ResNet-101 for correct class. \emph{ResNet-101 predicted class} is class predicted by ResNet-101 wrongly and \emph{ResNet-101 predicted class accuracy} is accuracy given in this class. \emph{FDResNet-101 accuracy} is accuracy of FDResNet-101 in the correct class, and \emph{FDResNet-101 accuracy in wrong class} is accuracy of the class wrongly predicted by ResNet-101.} 
\label{fig:model_predictions}
\end{figure*}

\section{Experimental Results for Image Classification}
In this section we present the experimental results on four benchmark datasets for image classification. We perform the rigorous experiments for classification and compare the results of the proposed FDResNet model with ResNet model. We also show the impact of sigma of Gaussian filtering and low vs high pass skip connections. The qualitative results are also presented to justify the working of the proposed model. The convergence is also shown in terms of the loss and accuracy plots.

\subsection{Quantitative Results}
The accuracy comparison for image classification on benchmark CIFAR-10, CIFAR-100, Caltech-256 and TinyImageNet datasets are summarized in Table \ref{tab:classification_results}. The results are computed as average over three independent trials. The standard deviation (sigma) of Gaussian filtering is set to 1 in the proposed FDResNet with varying combinations of kernel size for low pass and high pass skip connections. The results on CIFAR-10 and CIFAR-100 datasets are computed using ResNet50 and FDResNet50 models. Whereas, the results on Caltech-256 and TinyImageNet datasets are computed using ResNet101 and FDResNet101 models.
Following are the observations from the image classification results:
\begin{itemize}
    \item The proposed FDResNet model outperforms the ResNet model in most of the kernel settings of FDResNet.
    \item It is observed that the proposed FDResNet model with kernel size 3 for both low pass and high pass skip connections (L3, H3) leads to the best performance on CIFAR-10 and CIFAR-100 datasets. However, the kernel size (L5, H5) and (L7, H7) of FDResNet are the best performing on TinyImageNet and Caltech-256 datasets, respectively.
    \item The accuracy of the FDResNet50 with kernel size (L3, H3) are improved by $0.17 \%$ and $2.59 \%$ as compared to the ResNet50 on CIFAR-10 and CIFAR-100 datasets, respectively.
    \item The accuracy of the FDResNet101 model with kernel size (L7, H7) on Caltech-256 dataset and kernel size (L5, H5) on TinyImageNet dataset are improved by $7.99 \%$ and $9.25 \%$, respectively, as compared to the ResNet101 model.
    \item The results of kernel size (L3, H5) are poor than other settings, possibly due to the loss of information with higher kernel size for high pass filtering as compared to low pass filtering.
\end{itemize}

\begin{figure*}[!t]
    \centering
    \includegraphics[width=1\textwidth]{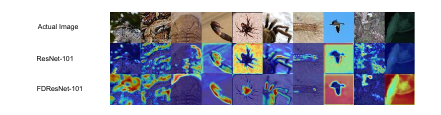}
    \caption{The saliency maps depicting the regions where the ResNet101 and FDResNet101 models focus on some sample images from TinyImageNet dataset. We use sigma value of 1 and kernel size of 5 for both low pass and high pass skip connections in FDResNet. These maps are produced at the very first convolution layer of the model.}
    \label{fig:saliency_map}
\end{figure*}

\subsection{Qualitative Results}
In order to justify the improved performance using the proposed model, the qualitative results on sample images of Caltech-256 and TinyImageNet datasets are depicted in Fig. \ref{fig:Caltech_Scores} and Fig. \ref{fig:Tiny_Scores}, respectively. The prediction scores using ResNet101 and FDResNet101 for the correct class are reported in Red and Green colors, respectively.
The FDResNet is used with a sigma value of 1 and kernel size of 3 on Caltech-256 dataset in Fig. \ref{fig:Caltech_Scores} and with sigma value of 1 and kernel size of 5 on TinyImageNet dataset in Fig. \ref{fig:Tiny_Scores}. It can be noted that ResNet101 fails in all of these examples as the class scores for the correct class (i.e., ResNet101 Accuracy) are very low and the class scores for the wrong predicted class (i.e., ResNet101 Predicted Class Accuracy for ResNet101 Predicted Class) are more. However, the FDResNet101 is able to classify these images to the correct class with high class scores. It shows the importance of the frequency disentangling in ResNet as different images have different perception of low and high frequency components.
 
In order to further observe the effect of frequency disentangling, we compute the saliency maps using the GradCAM \cite{gradcam} as illustrated in Fig. \ref{fig:saliency_map} using the ResNet101 and FDResNet101 models on some sample images of TinyImageNet dataset. The GradCAM helps to identify the prominent regions by using the gradients flowing through the network.
The FDResNet model in this experiment uses the filtering with unit sigma and 5 kernel size for both low pass and high pass skip connections. The $1^{st}$ convolution layer is used to visualize these saliency maps. The Red color represents higher importance while the Blue color represents lower importance. In these images we can see that the FDResNet model focuses on the representative regions in most of the cases even in the early layer of the network due to the better encoding of the low and high frequency information.


\begin{figure}[!t]
    \centering
    \includegraphics[width=0.49\columnwidth]{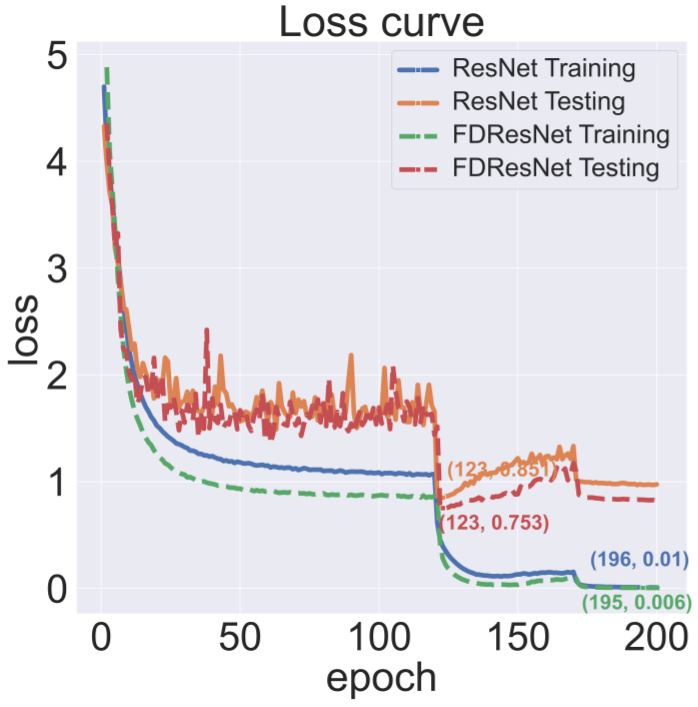}
     \includegraphics[width=0.49\columnwidth]{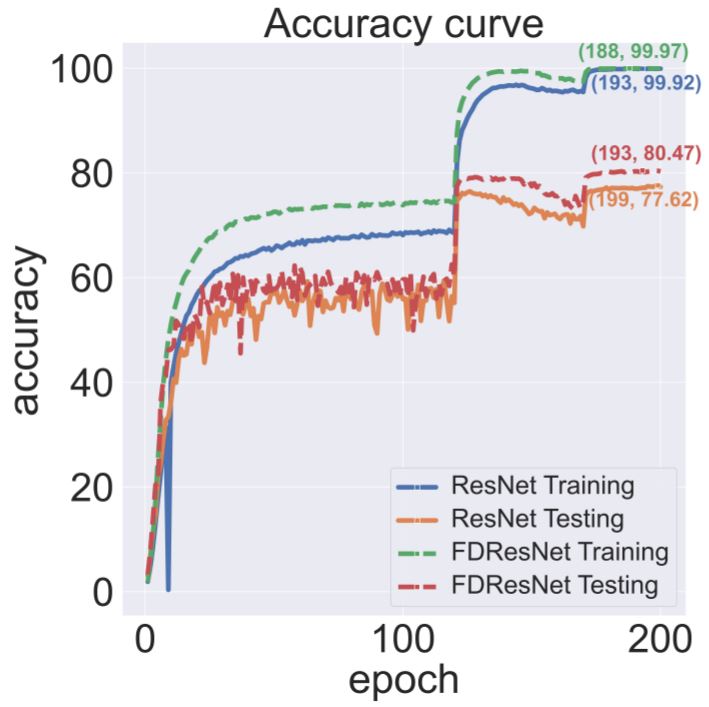}
    \caption{The convergence plots in terms of the loss (left) and accuracy (right) curves of training and testing of the ResNet50 and FDResNet50 on CIFAR-100 dataset. The FDResNet50 is used with settings of sigma value 1 and a kernel size 3 for both low pass and high pass skip connections.}
    \label{fig:convergence_cifar100}
\end{figure}

\begin{figure}[!t]
    \centering
    \includegraphics[width=0.49\columnwidth]{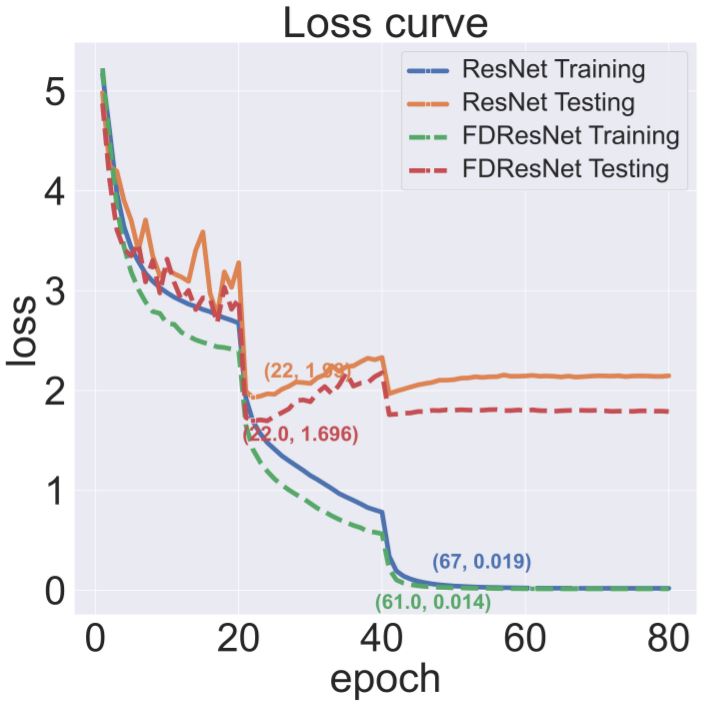}
    \includegraphics[width=0.49\columnwidth]{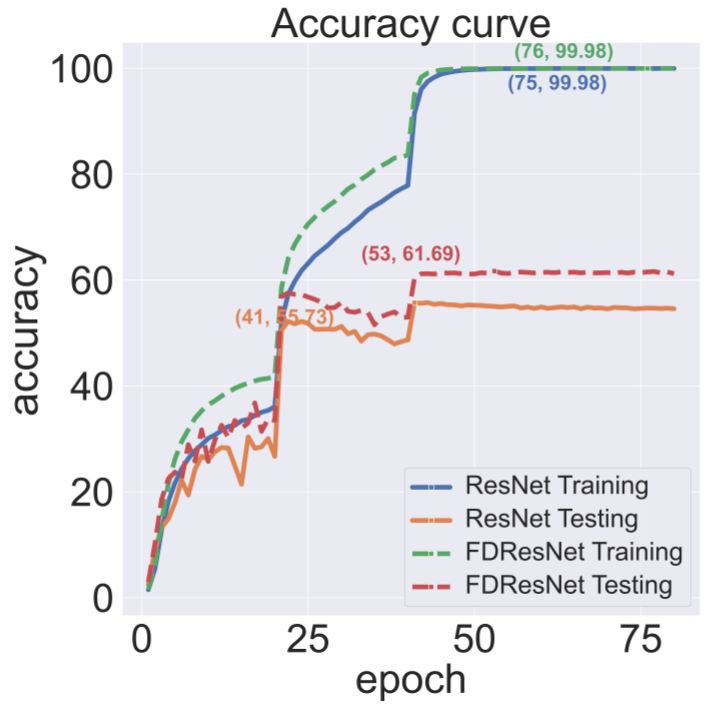}
    \caption{The convergence plots in terms of the loss (left) and accuracy (right) curves of training and testing of the ResNet101 and FDResNet101 on TinyImageNet dataset. The FDResNet101 is used with settings of sigma value 1 and kernel size 5 for both low pass and high pass skip connections.}
    \label{fig:convergence_tinyimagenet}
\end{figure}

\subsection{Convergence Analysis}
The loss and accuracy curves for each epoch are plotted to depict the convergence property in Fig. \ref{fig:convergence_cifar100} on CIFAR-100 dataset and Fig. \ref{fig:convergence_tinyimagenet} on TinyImageNet dataset. The training and testing plots are shown using ResNet50 and FDResNet50 models on CIFAR-100 dataset and using ResNet101 and FDResNet101 models on TinyImageNet dataset. The sigma value is set to 1 for both FDResNet50 and FDResNet101 models. Whereas, the kernel size is 3 for FDResNet50 and 5 for FDResNet101. It is asserted that the convergence using the proposed FDResNet models is better than the ResNet models. Moreover, the overfitting using the proposed models is also lower which indicates the better generalization capability of the FDResNet models.

\begin{table}[!t]
\caption{The performance of the proposed FDResNet model by considering only one skip connection. We consider only low pass or high pass skip connection with 3 and 5 kernel sizes. The sigma value is set as 1 for this experiment.}
\centering
\renewcommand{\arraystretch}{1.2}
\resizebox{\columnwidth}{!}{
\begin{tabular}{|p{0.16\columnwidth}|p{0.16\columnwidth}|p{0.16\columnwidth}|p{0.16\columnwidth}|p{0.16\columnwidth}|p{0.16\columnwidth}|}
\hline
\multirow{2}{*}{Dataset} & ResNet & \multicolumn{4}{c|}{FDResNet Accuracy using kernel with sigma 1} \\ 
 \cline{3-6} 
& Accuracy & (L3, Nil) & (L5, Nil) & (Nil, H3) & (Nil, H5) \\ \hline
CIFAR-10 &
  ${\textbf{95.16}\pm0.12}$ &
  ${94.24\pm0.30}$ &
  ${94.38\pm0.38}$ &
  ${93.21\pm0.25}$ &
  ${93.37\pm0.45}$ \\ \hline
CIFAR-100 &
  ${\textbf{78.42}\pm0.60}$ &
  ${78.12\pm0.23}$ &
  ${76.69\pm0.27}$ &
  ${74.73\pm0.24}$ &
  ${75.35\pm0.44}$ \\ \hline
Caltech-256 &
  ${\textbf{58.17}\pm0.73}$ &
  ${56.84\pm1.27}$ &
  ${54.57\pm0.74}$ &
  ${38.67\pm1.11}$ &
  ${41.57\pm1.79}$ \\ \hline
TinyImageNet &
  ${57.18\pm0.35}$ &
  ${\textbf{59.89}\pm0.58}$ &
  ${57.32\pm0.30}$ &
  ${36.98\pm0.83}$ &
  ${37.24\pm0.39}$ \\ \hline
\end{tabular}
}
\label{tab:single_skip}
\end{table}

\subsection{Impact of Low and High Pass Skip Connection}
The proposed FDResNet model considers two skip connections, namely low pass and high pass. In this experiment, we analyze the impact of individual skip connections, i.e., low pass and high pass skip connections. The classification accuracy using individual skip connection is detailed in Table \ref{tab:single_skip} using FDResNet with (L3, Nil), (L5, Nil), (Nil, H3), and (Nil, H5) skip connection settings, where Nil means corresponding skip connection is missing. The sigma value for filtering is set to 1. It is observed that the performance of FDResNet model with only one skip connection is not improved than the ResNet model. Moreover, the performance of the low pass skip connection is better than the high pass skip connection as high pass filtering incurs more information loss. However, when both low pass and high pass skip connections are used together then the performance of the proposed model is superior as shown in Table \ref{tab:classification_results}. It confirms that the utilization of different frequency information using frequency disentangling leads to better performance.

\begin{table}[!t]
\caption{The results using 0.5 sigma value by varying the kernel size for both skip connections of FDResNet on CIFAR-10, CIFAR-100, Caltech-256, and TinyImageNet datasets. The best result for a dataset is highlighted in bold.}
\centering
\renewcommand{\arraystretch}{1.2}
\resizebox{\columnwidth}{!}{
\begin{tabular}{|p{0.18\columnwidth}|p{0.18\columnwidth}|p{0.18\columnwidth}|p{0.18\columnwidth}|p{0.18\columnwidth}|}
\hline
\multirow{2}{*}{Dataset} & ResNet & \multicolumn{3}{c|}{FDResNet Accuracy using kernel with sigma 0.5} \\ 
 \cline{3-5} 
& Accuracy & (L3, H3) & (L5, H5) & (L7, H7) \\ \hline
CIFAR-10 & ${95.16\pm0.12}$ & ${95.33\pm0.24}$ & ${\textbf{95.39}\pm0.06}$ & ${95.19\pm0.10}$ \\ \hline
CIFAR-100 & ${78.42\pm0.60}$ & ${\textbf{80.08}\pm0.32}$ & ${79.58\pm0.33}$ & ${79.63\pm0.40}$ \\ \hline
Caltech-256 & ${58.17\pm0.73}$ & ${59.02\pm0.17}$ & ${59.25\pm1.29}$ & ${\textbf{60.64}\pm2.17}$ \\ \hline
TinyImageNet & ${57.18\pm0.35}$ & ${60.39\pm0.74}$ & ${60.18\pm0.56}$ & ${\textbf{60.46}\pm0.36}$ \\ \hline
\end{tabular}
}
\label{tab:sigma_half}
\end{table}

\begin{table}[!t]
\caption{The results using 1.5 sigma value by varying the kernel size for both skip connections of FDResNet on CIFAR-10, CIFAR-100, Caltech-256, and TinyImageNet datasets. The best result for a dataset is highlighted in bold.}
\centering
\renewcommand{\arraystretch}{1.3}
\resizebox{\columnwidth}{!}{
\begin{tabular}{|p{0.18\columnwidth}|p{0.18\columnwidth}|p{0.18\columnwidth}|p{0.18\columnwidth}|p{0.18\columnwidth}|}
\hline
\multirow{2}{*}{Dataset} & ResNet & \multicolumn{3}{c|}{FDResNet Accuracy using kernel with sigma 1.5} \\ 
 \cline{3-5} 
& Accuracy & (L3, H3) & (L5, H5) & (L7, H7) \\ \hline
CIFAR-10 & ${95.16\pm0.12}$ & ${\textbf{95.41}\pm0.13}$ & ${94.92\pm0.23}$ & ${95.02\pm0.11}$ \\ \hline
CIFAR-100 & ${78.42\pm0.60}$ & ${\textbf{80.18}\pm0.50}$ & ${79.69\pm0.18}$ & ${79.12\pm0.15}$ \\ \hline
Caltech-256 & ${58.17\pm0.73}$ & ${60.67\pm1.63}$ & ${61.14\pm0.92}$ & ${\textbf{61.57}\pm0.53}$ \\ \hline
TinyImageNet & ${57.18\pm0.35}$ & ${61.84\pm0.54}$ & ${\textbf{62.23}\pm0.32}$ & ${60.96\pm1.28}$ \\ \hline
\end{tabular}
}
\label{tab:sigma_oneandhalf}
\end{table}

\begin{table}[!t]
\caption{The results using trainable sigma by varying the kernel size for both skip connections of FDResNet on CIFAR-10, CIFAR-100, Caltech-256, and TinyImageNet datasets. The sigma parameter is initialized to 1. The best result for a dataset is highlighted in bold. }
\centering
\renewcommand{\arraystretch}{1.2}
\resizebox{\columnwidth}{!}{
\begin{tabular}{|p{0.18\columnwidth}|p{0.18\columnwidth}|p{0.18\columnwidth}|p{0.18\columnwidth}|p{0.18\columnwidth}|}
\hline
\multirow{2}{*}{Dataset} & ResNet & \multicolumn{3}{c|}{FDResNet Accuracy using kernel with trainable sigma} \\ 
 \cline{3-5} 
& Accuracy & (L3, H3) & (L5, H5) & (L7, H7) \\ \hline
CIFAR-10 & ${\textbf{95.16}\pm0.12}$ & ${94.02\pm1.03}$ & ${94.02\pm0.67}$ & ${93.70\pm0.51}$ \\ \hline
CIFAR-100 & ${\textbf{78.42}\pm0.60}$ & ${73.94\pm0.80}$ & ${73.83\pm0.36}$ & ${74.16\pm0.68}$ \\ \hline
Caltech-256 & ${58.17\pm0.73}$ & ${\textbf{61.46}\pm1.07}$ & ${60.53\pm0.79}$ & ${60.32\pm2.61}$ \\ \hline
TinyImageNet & ${57.18\pm0.35}$ & ${\textbf{61.89}\pm0.37}$ & ${61.28\pm0.30}$ & ${59.15\pm2.23}$ \\ \hline
\end{tabular}
}
\label{tab:sigma-trainable}
\end{table}

\subsection{Impact of Sigma of Filtering}
In the results shown in Table \ref{tab:classification_results}, we use the standard deviation (i.e., sigma) of the filter as 1 for both low pass and high pass in FDResNet. In order to observe the impact of the sigma value, we report the classification results by using the sigma value as 0.5 and 1.5 for both low pass and high pass filtering in Table \ref{tab:sigma_half} and Table \ref{tab:sigma_oneandhalf}, respectively. We also consider sigma as a trainable parameter with initialization to 1 and show the results in Table \ref{tab:sigma-trainable}. It can be perceived that the performance of the proposed FDResNet model is better than ResNet model in most of the cases even with different sigma values. The performance of the trainable sigma is comparatively lower in majority of the cases, possibly due to the overfitting caused by the trainable sigma. Thus, it validate our hypothesis that a fix sized filtering can improve the generalization. It is also noticed that the trainable sigma leads to very high sigma values near to 10 at the end of training, which is not the ideal case in filtering. 
On an average considering all four datasets, we observe a superior performance by the proposed FDResNet model using the sigma value as 1 as compared to 0.5, 1.5 and trainable sigma.

\begin{table*}[!t]
\caption{The performance comparison between the ResNet and the proposed FDResNet models on the filtered test images to depict the filtering robustness in CIFAR-10 and TinyImageNet datasets. Six test sets are created for each dataset including three sets of the high pass filtered images and three sets of low pass filtered images. The kernel sizes are 3, 5, and 7 with unit standard deviation. Low pass test images become blurred and high pass test images become noisy with prominent edge information.}
\centering
\renewcommand{\arraystretch}{1.2}
\begin{tabular}{|l|c|c|c|c|c|c|c|}
\hline
\multicolumn{1}{|c|}{\multirow{3}{*}{Dataset}} & \multirow{3}{*}{Filter} & \multicolumn{6}{c|}{Kernel width for filtering of test images}\\ \cline{3-8} 
& & \multicolumn{2}{c|}{3} & \multicolumn{2}{c|}{5} & \multicolumn{2}{c|}{7} \\ \cline{3-8} & & ResNet & FDResNet & ResNet & FDResNet & ResNet & FDResNet \\ \hline
\multirow{2}{*}{CIFAR-10} & Highpass & \textbf{27.56} & 11.41 & \textbf{27.38} & 11.42 & \textbf{27.32} & 11.42 \\ \cline{2-8} 
& Lowpass  & 49.96  & \textbf{72.22}    & 36.52  & \textbf{53.54}    & 35.59  & \textbf{54.19}    \\ \hline
\multirow{2}{*}{TinyImageNet} & Highpass & \textbf{1.99} & 1.29 & \textbf{2.01} & 1.32  & \textbf{2.01} & 1.32 \\ \cline{2-8} 
& Lowpass  & 24.86  & \textbf{28.07}    & 18.95  & \textbf{22.28}    & 18.54  & \textbf{22.13}    \\ \hline
\end{tabular}
\label{tab:robustness}
\end{table*}

\subsection{Filtering Robustness Analysis}
In order to depict the robustness property, we create multiple filtered test sets from the original test set and compute the accuracy on these filtered test sets. Six test sets are created for each dataset including three sets of high pass filtered images and three sets of low pass filtered images. The kernel sizes are 3, 5, and 7 with unit standard deviation. Low pass test images become blurred and high passed test images become noisy with prominent edge information. In this experiment, we do not perform the training, rather test the trained models on the filtered test sets. We use the FDResNet model with the settings which gave the best results for that particular dataset.
The results using trained ResNet and FDResNet models on the filtered test sets of CIFAR-10 and TinyImageNet datasets are summarized in Table \ref{tab:robustness}.
On both the CIFAR10 and TinyImageNet datasets, we spot that the proposed FDResNet model outperforms the ResNet counterparts by a large margin on low pass filtered test sets. However, both the models suffer on high pass filtered test sets as the input images lack  the sufficient information. As the FDResNet model disentangles the frequency information, the low pass disentangling on the high pass filtered input images leads to zero information, resulting in poor performance. Thus, the robustness of the proposed model is better on low pass filtered input images, but sensitive on high pass filtered input images.

\begin{table}[!t]
\caption{The image retrieval results in terms of the mean average precision (mAP) using the ResNet101 and FDResNet101 models on CIFAR-10, CIFAR-100 and TinyImageNet datasets. The FDResNet101 uses sigma value as 1 and kernel size as 3 for both filtering blocks. The superior result on a dataset is highlighted in bold.}
\centering
\renewcommand{\arraystretch}{1.3}
\resizebox{\columnwidth}{!}{
\begin{tabular}{|l|c|c|c|c|c|c|}
\hline
\multicolumn{1}{|c|}{\multirow{2}{*}{Model}} & \multicolumn{3}{c|}{mAP for Image Retrieval} \\  \cline{2-4}
& CIFAR-10 & CIFAR-100 & TinyImageNet \\\hline
ResNet101 & 0.863 & 0.608 & 0.422 \\\hline
FDResNet101 & \textbf{0.889} & \textbf{0.664} & \textbf{0.493} \\\hline
\end{tabular}
}
\label{tab:retrieval}
\end{table}

\section{Experimental Results for Image Retrieval}
In order to show the generalization ability of the proposed FDResNet model across different tasks, the image retrieval experiments are also performed.
Image retrieval attempts to find the alike images from a large image dataset for a given query image. The mean average precision (mAP) is the predominantly used criterion in the literature to judge the performance of the image retrieval methods \cite{dubey2021decade}. The precision is defined as the ratio of the number of correctly retrieved images with the total number retrieved images. 


We consider the feature vector of last layer for image retrieval. Table \ref{tab:retrieval} illustrated the image retrieval mAPs using the ResNet101 and the proposed FDResNet101 models. We compute the results using FDResNet using sigma value as 1 and kernel size as 3 for both low pass and high pass skip connections. In this experiment, we attain the improvement in mAPs using the proposed frequency disentangling idea. The mAP using the FDResNet101 is improved by $3.01 \%$, $9.21 \%$ and $16.82 \%$ as compared to the ResNet101 on CIFAR-10, CIFAR-100 and TinyImageNet datasets, respectively.

\section{Conclusion}
In this paper, a novel frequency disentangling based residual network (FDResNet) is proposed. The low pass and high pass filtering is exploited in order to incorporate the frequency disentangling property. The residual block in the proposed model consists of low pass and high pass skip connections. Thus, the proposed model better encodes the frequency information in the abstract feature and generalizes on unseen data.
The image classification results in terms of the accuracy using the proposed FDResNet model are better than the ResNet model on CIFAR-10, CIFAR-100, Caltech-256 and TinyImageNet datasets. It is observed that the proposed model better focuses on the salient high and low frequency regions in the image. The convergence analysis confirms the generalization capability of the proposed model. The robustness of the proposed model is observed against the low pass filtering on the input images. The performance of the proposed model is noticed as superior across different standard deviations of filtering.
We also notice the improved performance for image retrieval task using the features learnt by the proposed FDResNet model as compared to the ResNet model in terms of the mAPs.  
Thus, the proposed frequency disentangling idea leads to a better learning of robust features of low and high frequency information, leading to better generalization across different tasks. 
The future works can include the extension of the proposed frequency disentangling idea with other type of networks on different computer vision applications.



\section*{Acknowledgement}
This research is funded by the Global Innovation and Technology Alliance (GITA) on Behalf of Department of Science and Technology (DST), Govt. of India under India-Taiwan joint project with Project Code GITA/DST/TWN/P-83/2019.
We would like to acknowledge the NVIDIA for supporting with NVIDIA GPUs and Google Colab service for providing free computational resources which have been used in this research for the experiments.

{\small
\bibliographystyle{IEEEtran}
\bibliography{References}

\begin{thebibliography}{10}
\providecommand{\url}[1]{#1}
\csname url@samestyle\endcsname
\providecommand{\newblock}{\relax}
\providecommand{\bibinfo}[2]{#2}
\providecommand{\BIBentrySTDinterwordspacing}{\spaceskip=0pt\relax}
\providecommand{\BIBentryALTinterwordstretchfactor}{4}
\providecommand{\BIBentryALTinterwordspacing}{\spaceskip=\fontdimen2\font plus
\BIBentryALTinterwordstretchfactor\fontdimen3\font minus
  \fontdimen4\font\relax}
\providecommand{\BIBforeignlanguage}[2]{{%
\expandafter\ifx\csname l@#1\endcsname\relax
\typeout{** WARNING: IEEEtran.bst: No hyphenation pattern has been}%
\typeout{** loaded for the language `#1'. Using the pattern for}%
\typeout{** the default language instead.}%
\else
\language=\csname l@#1\endcsname
\fi
#2}}
\providecommand{\BIBdecl}{\relax}
\BIBdecl

\bibitem{alexnet}
A.~Krizhevsky, I.~Sutskever, and G.~E. Hinton, ``Imagenet classification with
  deep convolutional neural networks,'' in \emph{Advances in neural information
  processing systems}, 2012, pp. 1097--1105.

\bibitem{liu2017survey}
W.~Liu, Z.~Wang, X.~Liu, N.~Zeng, Y.~Liu, and F.~E. Alsaadi, ``A survey of deep
  neural network architectures and their applications,'' \emph{Neurocomputing},
  vol. 234, pp. 11--26, 2017.

\bibitem{lecun2015deep}
Y.~LeCun, Y.~Bengio, and G.~Hinton, ``Deep learning,'' \emph{nature}, vol. 521,
  no. 7553, pp. 436--444, 2015.

\bibitem{googlenet}
C.~Szegedy, W.~Liu, Y.~Jia, P.~Sermanet, S.~Reed, D.~Anguelov, D.~Erhan,
  V.~Vanhoucke, and A.~Rabinovich, ``Going deeper with convolutions,'' in
  \emph{IEEE Conference on Computer Vision and Pattern Recognition}, 2015, pp.
  1--9.

\bibitem{resnet}
K.~He, X.~Zhang, S.~Ren, and J.~Sun, ``Deep residual learning for image
  recognition,'' in \emph{IEEE Conference on Computer Vision and Pattern
  Recognition}, 2016, pp. 770--778.

\bibitem{iold}
S.~R. Dubey, S.~K. Singh, and R.~K. Singh, ``Rotation and illumination
  invariant interleaved intensity order-based local descriptor,'' \emph{IEEE
  Transactions on Image Processing}, vol.~23, no.~12, pp. 5323--5333, 2014.

\bibitem{lwp}
------, ``Local wavelet pattern: a new feature descriptor for image retrieval
  in medical ct databases,'' \emph{IEEE Transactions on Image Processing},
  vol.~24, no.~12, pp. 5892--5903, 2015.

\bibitem{mdlbp}
------, ``Multichannel decoded local binary patterns for content-based image
  retrieval,'' \emph{IEEE Transactions on Image Processing}, vol.~25, no.~9,
  pp. 4018--4032, 2016.

\bibitem{schmidhuber2015deep}
J.~Schmidhuber, ``Deep learning in neural networks: An overview,'' \emph{Neural
  networks}, vol.~61, pp. 85--117, 2015.

\bibitem{liu2021scene}
S.~Liu, G.~Tian, Y.~Zhang, and P.~Duan, ``Scene recognition mechanism for
  service robot adapting various families: A cnn-based approach using
  multi-type cameras,'' \emph{IEEE Transactions on Multimedia}, 2021.

\bibitem{basha2020impact}
S.~S. Basha, S.~R. Dubey, V.~Pulabaigari, and S.~Mukherjee, ``Impact of fully
  connected layers on performance of convolutional neural networks for image
  classification,'' \emph{Neurocomputing}, vol. 378, pp. 112--119, 2020.

\bibitem{dubey2019diffgrad}
S.~R. Dubey, S.~Chakraborty, S.~K. Roy, S.~Mukherjee, S.~K. Singh, and B.~B.
  Chaudhuri, ``Diffgrad: an optimization method for convolutional neural
  networks,'' \emph{IEEE Transactions on Neural Networks and Learning Systems},
  2019.

\bibitem{ren2015faster}
S.~Ren, K.~He, R.~Girshick, and J.~Sun, ``Faster r-cnn: Towards real-time
  object detection with region proposal networks,'' in \emph{Advances in neural
  information processing systems}, 2015, pp. 91--99.

\bibitem{he2017mask}
K.~He, G.~Gkioxari, P.~Doll{\'a}r, and R.~Girshick, ``Mask r-cnn,'' in
  \emph{IEEE International Conference on Computer Vision}, 2017, pp.
  2961--2969.

\bibitem{ma2019iwave}
H.~Ma, D.~Liu, R.~Xiong, and F.~Wu, ``iwave: Cnn-based wavelet-like transform
  for image compression,'' \emph{IEEE Transactions on Multimedia}, vol.~22,
  no.~7, pp. 1667--1679, 2019.

\bibitem{jin2019flexible}
Z.~Jin, M.~Z. Iqbal, D.~Bobkov, W.~Zou, X.~Li, and E.~Steinbach, ``A flexible
  deep cnn framework for image restoration,'' \emph{IEEE Transactions on
  Multimedia}, vol.~22, no.~4, pp. 1055--1068, 2019.

\bibitem{schroff2015facenet}
F.~Schroff, D.~Kalenichenko, and J.~Philbin, ``Facenet: A unified embedding for
  face recognition and clustering,'' in \emph{IEEE Conference on Computer
  Vision and Pattern Recognition}, 2015, pp. 815--823.

\bibitem{srivastava2019performance}
Y.~Srivastava, V.~Murali, and S.~R. Dubey, ``A performance comparison of loss
  functions for deep face recognition,'' in \emph{Seventh National Conference
  on Computer Vision, Pattern Recognition, Image Processing and Graphics},
  2019.

\bibitem{tajrobehkar2021align}
M.~Tajrobehkar, K.~Tang, H.~Zhang, and J.~H. Lim, ``Align r-cnn: A pairwise
  head network for visual relationship detection,'' \emph{IEEE Transactions on
  Multimedia}, 2021.

\bibitem{choi2020combining}
J.~Y. Choi and B.~Lee, ``Combining of multiple deep networks via ensemble
  generalization loss, based on mri images, for alzheimer's disease
  classification,'' \emph{IEEE Signal Processing Letters}, vol.~27, pp.
  206--210, 2020.

\bibitem{lbpdad}
S.~R. Dubey, S.~K. Roy, S.~Chakraborty, S.~Mukherjee, and B.~B. Chaudhuri,
  ``Local bit-plane decoded convolutional neural network features for
  biomedical image retrieval,'' \emph{Neural Computing and Applications}, pp.
  1--13, 2019.

\bibitem{tian2020coarse}
C.~Tian, Y.~Xu, W.~Zuo, B.~Zhang, L.~Fei, and C.-W. Lin, ``Coarse-to-fine cnn
  for image super-resolution,'' \emph{IEEE Transactions on Multimedia},
  vol.~23, pp. 1489--1502, 2020.

\bibitem{que2020attentive}
Y.~Que, S.~Li, and H.~J. Lee, ``Attentive composite residual network for robust
  rain removal from single images,'' \emph{IEEE Transactions on Multimedia},
  2020.

\bibitem{park2021dynamic}
K.~Park, J.~W. Soh, and N.~I. Cho, ``Dynamic residual self-attention network
  for lightweight single image super-resolution,'' \emph{IEEE Transactions on
  Multimedia}, 2021.

\bibitem{akbari2021learned}
M.~Akbari, J.~Liang, J.~Han, and C.~Tu, ``Learned multi-resolution
  variable-rate image compression with octave-based residual blocks,''
  \emph{IEEE Transactions on Multimedia}, 2021.

\bibitem{chen2020reverse}
S.~Chen, X.~Tan, B.~Wang, H.~Lu, X.~Hu, and Y.~Fu, ``Reverse attention-based
  residual network for salient object detection,'' \emph{IEEE Transactions on
  Image Processing}, vol.~29, pp. 3763--3776, 2020.

\bibitem{zhou2020salient}
W.~Zhou, J.~Wu, J.~Lei, J.-N. Hwang, and L.~Yu, ``Salient object detection in
  stereoscopic 3d images using a deep convolutional residual autoencoder,''
  \emph{IEEE Transactions on Multimedia}, 2020.

\bibitem{liu2020real}
S.~Liu, K.-H. Thung, W.~Lin, P.-T. Yap, and D.~Shen, ``Real-time quality
  assessment of pediatric mri via semi-supervised deep nonlocal residual neural
  networks,'' \emph{IEEE Transactions on Image Processing}, vol.~29, pp.
  7697--7706, 2020.

\bibitem{tang2020br}
C.~Tang, X.~Liu, S.~An, and P.~Wang, ``Br2net: Defocus blur detection via a
  bidirectional channel attention residual refining network,'' \emph{IEEE
  Transactions on Multimedia}, vol.~23, pp. 624--635, 2020.

\bibitem{yeh2019multi}
C.-H. Yeh, C.-H. Huang, and L.-W. Kang, ``Multi-scale deep residual
  learning-based single image haze removal via image decomposition,''
  \emph{IEEE Transactions on Image Processing}, vol.~29, pp. 3153--3167, 2019.

\bibitem{fan2017label}
Y.-Y. Fan, S.~Liu, B.~Li, Z.~Guo, A.~Samal, J.~Wan, and S.~Z. Li, ``Label
  distribution-based facial attractiveness computation by deep residual
  learning,'' \emph{IEEE Transactions on Multimedia}, vol.~20, no.~8, pp.
  2196--2208, 2017.

\bibitem{batchnorm}
S.~Ioffe and C.~Szegedy, ``Batch normalization: Accelerating deep network
  training by reducing internal covariate shift,'' in \emph{International
  Conference on Machine Learning}, 2015, pp. 448--456.

\bibitem{perez2017effectiveness}
L.~Perez and J.~Wang, ``The effectiveness of data augmentation in image
  classification using deep learning,'' \emph{arXiv preprint arXiv:1712.04621},
  2017.

\bibitem{dropout}
N.~Srivastava, G.~Hinton, A.~Krizhevsky, I.~Sutskever, and R.~Salakhutdinov,
  ``Dropout: a simple way to prevent neural networks from overfitting,''
  \emph{The journal of machine learning research}, vol.~15, no.~1, pp.
  1929--1958, 2014.

\bibitem{agrawal2020using}
A.~Agrawal and N.~Mittal, ``Using cnn for facial expression recognition: a
  study of the effects of kernel size and number of filters on accuracy,''
  \emph{The Visual Computer}, vol.~36, no.~2, pp. 405--412, 2020.

\bibitem{hayou2018selection}
S.~Hayou, A.~Doucet, and J.~Rousseau, ``On the selection of initialization and
  activation function for deep neural networks,'' \emph{arXiv preprint
  arXiv:1805.08266}, 2018.

\bibitem{yedla2021performance}
R.~R. Yedla and S.~R. Dubey, ``On the performance of convolutional neural
  networks under high and low frequency information,'' in \emph{International
  Conference on Computer Vision and Image Processing}, 2021, p. 214.

\bibitem{vuilleumier2003distinct}
P.~Vuilleumier, J.~L. Armony, J.~Driver, and R.~J. Dolan, ``Distinct spatial
  frequency sensitivities for processing faces and emotional expressions,''
  \emph{Nature neuroscience}, vol.~6, no.~6, pp. 624--631, 2003.

\bibitem{monson2014perceptual}
B.~B. Monson, E.~J. Hunter, A.~J. Lotto, and B.~H. Story, ``The perceptual
  significance of high-frequency energy in the human voice,'' \emph{Frontiers
  in psychology}, vol.~5, p. 587, 2014.

\bibitem{yi2020bsd}
Z.~Yi, Z.~Chen, H.~Cai, W.~Mao, M.~Gong, and H.~Zhang, ``Bsd-gan: Branched
  generative adversarial network for scale-disentangled representation learning
  and image synthesis,'' \emph{IEEE Transactions on Image Processing}, vol.~29,
  pp. 9073--9083, 2020.

\bibitem{chen2020did}
X.~Chen, Y.~Wang, J.~Liu, and Y.~Qiao, ``Did:
  Disentangling-imprinting-distilling for continuous low-shot detection,''
  \emph{IEEE Transactions on Image Processing}, vol.~29, pp. 7765--7778, 2020.

\bibitem{chen2020rgbd}
H.~Chen, Y.~Deng, Y.~Li, T.-Y. Hung, and G.~Lin, ``Rgbd salient object
  detection via disentangled cross-modal fusion,'' \emph{IEEE Transactions on
  Image Processing}, vol.~29, pp. 8407--8416, 2020.

\bibitem{kottayil2017blind}
N.~K. Kottayil, G.~Valenzise, F.~Dufaux, and I.~Cheng, ``Blind quality
  estimation by disentangling perceptual and noisy features in high dynamic
  range images,'' \emph{IEEE Transactions on Image Processing}, vol.~27, no.~3,
  pp. 1512--1525, 2017.

\bibitem{imagenet}
O.~Russakovsky, J.~Deng, H.~Su, J.~Krause, S.~Satheesh, S.~Ma, Z.~Huang,
  A.~Karpathy, A.~Khosla, M.~Bernstein \emph{et~al.}, ``Imagenet large scale
  visual recognition challenge,'' \emph{International journal of computer
  vision}, vol. 115, no.~3, pp. 211--252, 2015.

\bibitem{vggnet}
K.~Simonyan and A.~Zisserman, ``Very deep convolutional networks for
  large-scale image recognition,'' \emph{arXiv preprint arXiv:1409.1556}, 2014.

\bibitem{densenet}
G.~{Huang}, Z.~{Liu}, L.~{Van Der Maaten}, and K.~Q. {Weinberger}, ``Densely
  connected convolutional networks,'' pp. 2261--2269, 2017.

\bibitem{resnext}
S.~{Xie}, R.~{Girshick}, P.~{Dollár}, Z.~{Tu}, and K.~{He}, ``Aggregated
  residual transformations for deep neural networks,'' pp. 5987--5995, 2017.

\bibitem{resnetstocasticdepth}
G.~Huang, Y.~Sun, Z.~Liu, D.~Sedra, and K.~Weinberger, ``Deep networks with
  stochastic depth,'' vol. 9908, pp. 646--661, 10 2016.

\bibitem{senet}
J.~Hu, L.~Shen, and G.~Sun, ``Squeeze-and-excitation networks,'' in \emph{IEEE
  Conference on Computer Vision and Pattern Recognition}, 2018, pp. 7132--7141.

\bibitem{krizhevsky2009learning}
A.~Krizhevsky, ``Learning multiple layers of features from tiny images,''
  \emph{Master's thesis, University of Tront}, 2009.

\bibitem{caltech256}
G.~Griffin, A.~Holub, and P.~Perona, ``Caltech-256 object category dataset,''
  \emph{CalTech Report}, 03 2007.

\bibitem{le2015tiny}
Y.~Le and X.~Yang, ``Tiny imagenet visual recognition challenge,'' \emph{CS
  231N}, vol.~7, 2015.

\bibitem{gradcam}
R.~R. Selvaraju, M.~Cogswell, A.~Das, R.~Vedantam, D.~Parikh, and D.~Batra,
  ``Grad-cam: Visual explanations from deep networks via gradient-based
  localization,'' in \emph{IEEE International Conference on Computer Vision},
  2017, pp. 618--626.

\bibitem{dubey2021decade}
S.~R. Dubey, ``A decade survey of content based image retrieval using deep
  learning,'' \emph{IEEE Transactions on Circuits and Systems for Video
  Technology}, 2021.

\end{thebibliography}
}

\end{document}